\documentclass[letterpaper]{article} % DO NOT CHANGE THIS
\usepackage[submission]{aaai23}  % DO NOT CHANGE THIS
\usepackage{times}  % DO NOT CHANGE THIS
\usepackage{helvet}  % DO NOT CHANGE THIS
\usepackage{courier}  % DO NOT CHANGE THIS
\usepackage[hyphens]{url}  % DO NOT CHANGE THIS
\usepackage{graphicx} % DO NOT CHANGE THIS
\usepackage{natbib}  % DO NOT CHANGE THIS AND DO NOT ADD ANY OPTIONS TO IT
\usepackage{caption} % DO NOT CHANGE THIS AND DO NOT ADD ANY OPTIONS TO IT
\usepackage{algorithm}
\usepackage{algorithmic}
\usepackage{booktabs}
\usepackage{multirow}
\usepackage{caption}
\usepackage{subcaption}
\usepackage{amsmath}
\usepackage{amssymb}
\usepackage{newfloat}
\usepackage{listings}
\DeclareCaptionStyle{ruled}{labelfont=normalfont,labelsep=colon,strut=off} % DO NOT CHANGE THIS
\floatstyle{ruled}
\newfloat{listing}{tb}{lst}{}
\floatname{listing}{Listing}
\title{Query-guided Attention in Vision Transformers for\\ Localizing Objects Using a Single Sketch}
\author{
Aditay Tripathi\textsuperscript{1} \hspace{1cm} Anand Mishra\textsuperscript{2} \hspace{1cm} Anirban Chakraborty\textsuperscript{1} \\
 \textsuperscript{1} Indian Institute of Science, Bengaluru\hspace{1cm} \textsuperscript{2} Indian Institute of Technology, Jodhpur\\
{\tt\small aditayt@iisc.ac.in}
}
\begin{document}

\maketitle

\begin{abstract}
In this work, we investigate the problem of sketch-based object localization on natural images, where given a crude hand-drawn sketch of an object, the goal is to localize all the instances of the same object on the target image. This problem proves difficult due to the abstract nature of hand-drawn sketches, variations in the style and quality of sketches, and the large domain gap existing between the sketches and the natural images. To mitigate these challenges, existing works proposed attention-based frameworks to incorporate query information into the image features. However, in these works, the query features are incorporated after the image features have already been independently learned, leading to inadequate alignment. In contrast, we propose a sketch-guided vision transformer encoder that uses cross-attention after each block of the transformer-based image encoder to learn query-conditioned image features leading to stronger alignment with the query sketch. Further, at the output of the decoder, the object and the sketch features are refined to bring the representation of relevant objects closer to the sketch query and thereby improve the localization. The proposed model also generalizes to the object categories not seen during training, as the target image features learned by our method are query-aware. Our localization framework can also utilize multiple sketch queries via a trainable novel sketch fusion strategy. The model is evaluated on the images from the public object detection benchmark, namely MS-COCO, using the sketch queries from QuickDraw! and Sketchy datasets. Compared with existing localization methods, the proposed approach gives a $6.6\%$ and $8.0\%$ improvement in mAP for seen objects using sketch queries from QuickDraw! and Sketchy datasets, respectively, and a $12.2\%$ improvement in AP@50 for large objects that are `unseen' during training.
\end{abstract}
\begin{figure}[!t]
    \centering
    \includegraphics[width=0.45\textwidth]{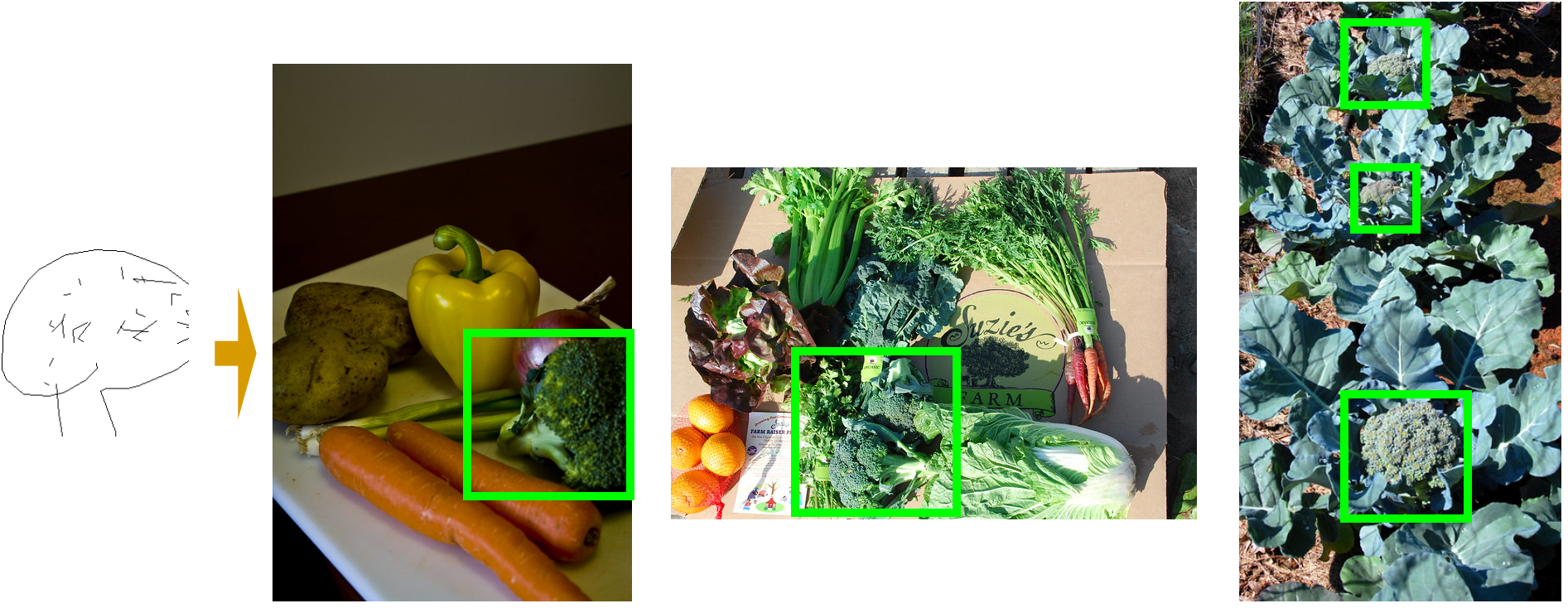}
    \caption{\label{fig:teaser}\textbf{Sketch-based object localization:} Consider a scenario where users wish to localize all the instances of the object \emph{broccoli} on a set of natural images, and (i) \emph{broccoli} is never seen during training, (ii) even at the inference time users do not have an image of \emph{broccoli} that can be used as a query, and (iii) the category name (``broccoli") is also unknown to the user. However, they prefer to draw a hand-drawn sketch of \emph{broccoli} to localize it on images. This is the `sketch-based object localization' problem. This paper significantly improves the performance of this recently introduced challenging task. \textbf{[Best viewed in color].}}
\end{figure}

\section{Introduction}
Detecting objects in a natural image is an exciting area of research in computer vision. It has seen tremendous progress over the past decade, partly thanks to evolving deep learning architectures~\cite{Song2021ViDTAE, Carion2020EndtoEndOD, Ren2015FasterRT}. However, the success of modern object detectors is often limited to localizing the object categories seen during training. In many practical scenarios, there is a need to have an object localization technique that can generalize well to unseen object categories; in other words, perform \emph{open-world object localization}. One of the directions in this space is to localize the objects on the natural scene guided by an image of an object as a query~\cite{Hsieh2019OneShotOD}. However, the availability of images for an object might be limited in many practical applications due to copyright, privacy concerns, and data collection overhead, especially for uncommon (often not natural) objects. Such a situation may also arise where users do not have access to query images of the object they wish to localize and also do not know what it is called but prefer to describe it via a hand-drawn sketch of it. Please refer to Figure~\ref{fig:teaser} to understand our motivation and objective.

In order to perform localization in the aforementioned scenario, ~\cite{Tripathi2020SketchGuidedOL} introduced the problem of sketch-based object localization, where the objective is to localize all the instances of the object on the natural image that corresponds to the object in the sketch query. This problem is significantly challenging primarily because of the abstract nature or `crudeness' of the sketches, the quality and variation in the style of the hand-drawn sketches produced by a diverse set of non-expert users, and the domain gap existing between the query sketches and the target natural images. As a first attempt to address some of these challenges, \cite{Tripathi2020SketchGuidedOL} proposed a cross-modal attention-based localization framework where they use a novel attention scheme to generate region proposals semantically relevant to the query sketch and then score them to obtain the most precise localization of the corresponding object of interest. However, being a proposal-based scheme, their framework is limited by the quality of the generated proposals, which is often poor for occluded or underrepresented objects. Moreover, the simplicity of the attention scheme proposed in this work also limits the localization performance. Giving a new dimension to this problem, more recently, Sketch-DETR has been proposed by~\cite{Riba2021LocalizingIF}. It extends popular detection transformer (DETR) object detector~\cite{Carion2020EndtoEndOD} to propose the state-of-the-art for sketch-based object localization task. Their proposed Sketch-DETR is an encoder-decoder transformer model that takes the target image and the sketch features at the input and uses multi-headed self-attention to refine their features. However, this refinement happens after the image features have already been learned, which leads to poor alignment between the image features and sketch features. 

Addressing the shortcomings in the existing approaches, we propose a novel sketch-guided vision transformer encoder based on the vision and detection transformer (ViDT)~\cite{Song2021ViDTAE} as the backbone. It should be noted that vanilla ViDT is query agnostic and learns image features independent of the query. Our proposed transformer learns the representation of the target image conditioned on the query sketch. After each block of the image encoder, multi-headed cross-attention is used, where the computed attention score between the image and the sketch features is used to fuse the sketch features into the image features. The target image features, thus obtained, are better aligned with the query sketch, leading to better query-guided localization performance. Additionally, a more fine-grained object-level fusion at the output of the decoder can lead to performance improvement. To this end, for each object, attention is used to incorporate sketch features and vice-versa. This helps bring the representation of relevant objects closer to the sketch query, enabling  precise localization.  
 
One key facet of our proposed framework is that it works respectably well under the challenging `open-world' setting, i.e., it can produce very accurate localization even for object categories unseen during training because the representation learned for the target image is aligned with the query sketch. Moreover, we extend our localization framework to utilize multiple sketches as queries via a trainable novel sketch fusion strategy that combines the complementary information present in multiple sketches to construct a complete representation of the object leading to better localization.

In summary, we make the following contributions: (i) We propose the novel \textit{sketch-guided vision transformer encoder} that learns the representation of the target image conditioned on the query sketch, which leads to better alignment between the image and the sketch features. This results in much improved query-guided localization. (ii) Additionally, we propose an object feature refinement strategy at the output of the decoder that utilizes attention to bring the features of the relevant objects closer to the sketch query, thereby further improving the localization performance. (iii) Our proposed approach achieves a substantial gain of~8\% over the best-reported results for sketch-based object localization on images from the MS-COCO dataset and query sketches from the Sketchy dataset. It, therefore, establishes a new state of the art for this task.

\section{Related Work}
\subsection{Object detection}
Object detection is a well-studied yet open area of research in computer vision. Broadly, object detection approaches can be grouped into (i) proposal-based and (ii) proposal-free methods. Although proposal-based methods~\cite{Uijlings2013SelectiveSF,Girshick2014RichFH} have several advantages, their performance is often limited by the quality of proposals it generates, which are often weak for occluded objects as well as `unseen' object categories. Improving proposal generation for unseen objects is an open area of research~\cite{accv}. 
Our work falls under proposal-free methods. Among proposal-free methods, the modern transformer-based object detectors~\cite{Zhu2021DeformableDD,Carion2020EndtoEndOD, Song2021ViDTAE} are state-of-the-art for object detection. These methods often have encoder-decoder models and utilize a fixed set of $[\mathtt{DET}]$ tokens to learn to localize and classify the objects in the image. Methods such as ViDT~\cite{Song2021ViDTAE} have made progress toward an encoder-free object detector, leading to fewer parameters and faster inference. These object detection methods are reasonably successful in the close-world setting. There is also a growing interest to address open-world object localization~\cite{owdetr,Hsieh2019OneShotOD,ovod}. In this space,~\cite{Hsieh2019OneShotOD} and~\cite{ovod} have proposed object localization in the one-shot setting and use the image of an object as a query. We address the problem of object localization problem for the scenario where the object category name is unknown and the query image for the object of interest is unavailable; rather, a crude sketch representation of the object is available for the query to perform one-shot object localization. This recently introduced problem is referred to as \emph{sketch-based object localization}. In the following section, we discuss the differences between our approach and existing methods in this space. 
\subsection{Sketch-based Object Localization}
Hand-drawn sketches have been applied to various computer vision tasks, examples include, sketch generation~\cite{DBLP:journals/ijcv/QiSWYPS22}, 3D reconstruction~\cite{DBLP:conf/3dim/LunGKMW17}, image and video retrieval~\cite{Radenovi2018DeepSM,Xu2021FineGrainedIS}, and retrieval of 3D shapes~\cite{Wang2015Sketchbased3S,Qi2021TowardFS}.
Recently,~\cite{Tripathi2020SketchGuidedOL} introduced the problem of sketch-based object localization in natural images. In this problem, given a sketch query, the task is to localize the corresponding objects in the target images. They proposed a model based on Faster-RCNN and a cross-attention module to generate region proposals relevant to the query. However, their model utilizes an inadequate attention mechanism that leads to subpar performance. More recently, \cite{Riba2021LocalizingIF} proposed a transformer-based approach namely Sketch-DETR. In this work, they concatenate the flattened sketch and image feature before feeding them through the DETR encoder. Although more expressive than the cross-modal attention mechanism, they incorporate the sketch information at the encoder where target image features have already been learned. In this work, we propose \emph{sketch-guided vision transformer encoder} that learns the representation of the target image conditioned on the query sketch by fusing the query information into the target image after each block of the transformer-based image encoder.

\begin{figure*}
    \centering
    \includegraphics[width=\textwidth]{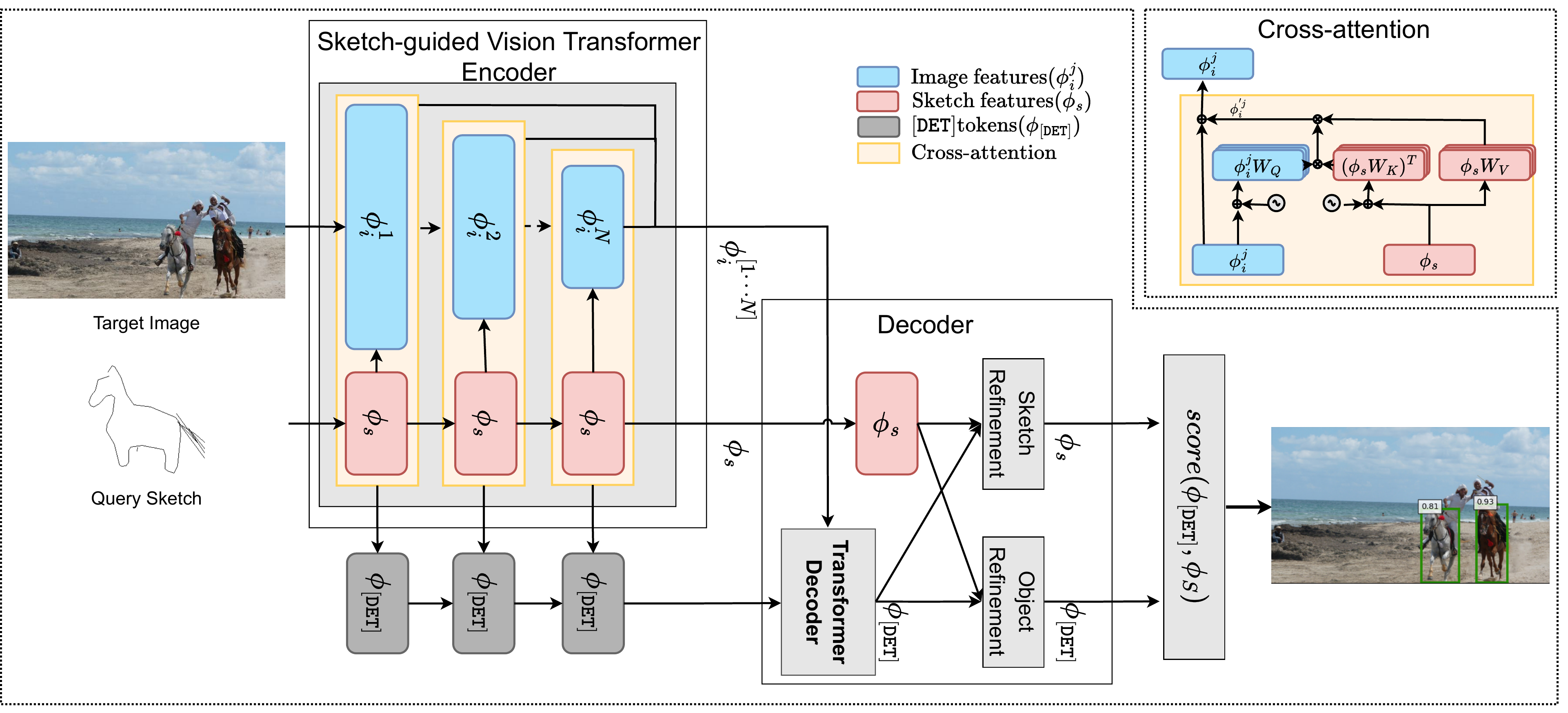}
    \caption{\label{fig:archi}The proposed sketch-guided object localization model consists of two main components: (a) sketch-guided vision transformer encoder (refer to Section~\ref{sec:s2i}) and (b) object and query refinement Decoder (refer to Section~\ref{sec:prop_scoring}). The sketch-guided transformer encoder takes target image at the input and generate sketch-conditioned features for it by fusing the sketch features into the target image after each block of the image encoder using \textbf{cross-attention} (shown in detail in the top-right). After getting object-level features at the output of the \textbf{transformer decoder} ($\phi_{[\mathtt{DET}]}$), the object features and the query sketch features are further refined to bring the features of the relevant object closer to the query sketch leading to better localization score. [Best viewed in color]. }
\end{figure*}

\section{Methodology}
\subsection{Task Definition and Proposed Architecture} 
Consider a dataset ${\cal D}=\{I,S\}$ where $I$ and $S$ are a set of natural images and hand-drawn sketches, respectively. Let $C$ be the set of all object categories present in ${\cal D}$. Each image $I_i \in I$ contains  bounding box annotations $B_i = \{(b_j, c_j)\}_{j=1}^{n_i}$ corresponding to all object instances (any of the $C$ categories) present in it. Here, $b_j$ is a rectangular box tightly surrounding the $j^{th}$ object instance and $c_j \in C$ is the category of that object. Given a sketch $s_k \in S$ and an image $I_i \in I$, the problem of sketch-based object localization involves localizing all the instances of the object in the image that correspond to the sketch $s_k$. We address this problem in both close-world, i.e., when
object category of sketch query is `seen' during training and open-world, i.e. when any example of the object category of sketch query is `unseen' during training.
A plausible approach to address sketch-based object localization is to extend the vision and detection transformer (ViDT)~\cite{Song2021ViDTAE} to support sketch queries. To this end, one can compare the representation of objects at the output of the decoder, given by $[\mathtt{DET}]$ tokens of ViDT, to the representation of the query sketch to obtain the best localization of the corresponding object. However, as the object and the query representations are learned independently in this trivial extension, it may lead to sub-optimal localization performance. In this work, we, therefore, propose a novel \emph{sketch-guided vision transformer encoder} that learns the representation of the target image conditioned on the query sketch. The target image representation learned this way is better aligned with the query sketch, leading to better localization performance. Moreover, the object and the query features are further refined at the output of the decoder to bring the object features, corresponding to ground truth, closer to the sketch query for better scoring. The proposed model is end-to-end trainable, and it is described in the following three sections: (i) Sketch-guided vision transformer encoder, (ii) object and query representation refinement, and (iii) scoring. The overall architecture of the proposed model is illustrated in Figure~\ref{fig:archi}.

\subsubsection{(i) Sketch-guided Vision Transformer Encoder:}
\label{sec:s2i}
Traditional image encoders such as ResNet~\cite{DBLP:journals/corr/HeZRS15} as well as modern transformer-based image encoder such as Swin~\cite{DBLP:conf/iccv/LiuL00W0LG21} have several layers of neural networks which are grouped into blocks and, generally an image is passed through all the blocks of the image encoder to learn its representation. Similar architectures can be used to learn sketch embeddings separately. However, independently learned representations are poorly aligned because of the large domain gap existing between the target natural image and the query sketch. We mitigate this issue by proposing to learn the representation of the target image conditioned on the query sketch by fusing the features of the sketch query with the image features at the output of each block of the transformer-based image encoder. In this work, we use the Swin transformer as an image encoder. Firstly, the representation of the sketch query $s$ is learned by passing it through a sketch encoder, and it is represented as $\phi_s \in \mathbb{R}^{d\times w\times h }$. Similarly, for the target image $I_i$, the representation at the output of the first block of the image encoder is given by $\phi_i^1 \in \mathbb{R}^{d \times w_{I} \times h_{I}}$.

A pooling layer follows each block in the image encoder. Therefore, the output of each block represents features of the target image at different scales. In this work, we utilize multi-headed cross-attention to fuse the sketch features into the image feature at different granularity of the image input. The image and the sketch features are first flattened before passing them through the attention module. The flattened image representation $\phi_i^{1}\in \mathbb{R}^{ w_{I}h_{I}\times d}$ is used as queries and the flattened sketch representation $\phi_s\in \mathbb{R}^{wh \times d}$ is used as key and value in the multi-headed cross-attention module. A 2D sinusoidal position embeddings are also added to the query and the key features to provide spatial information while calculating the attention weights. For the sake of brevity, here we show the attention calculation for a single head. However, in our module, we use multiple heads to learn the correspondence between the target image and the query sketch. The representation of the target image is updated as follows:

\begin{equation}
\label{eq:soft}
    \phi_i^{'1} = softmax\left(\frac{(\phi_i^1 \mathbf{W}_{Q_{1}})\left(\phi_s \mathbf{W}_{K_{1}}\right)^T}{\sqrt{d^{'}}}\right)\phi_s\mathbf{W}_{V_{1}},
\end{equation}
where $d^{'}$ is the dimension of the key vectors, $\mathbf{W}_{Q_{1}}, \mathbf{W}_{K_{1}} \in \mathbb{R}^{d\times d^{'}}$ and $ \mathbf{W}_{V_{1}}\in \mathbb{R}^{d \times d}$ is the projection matrices for the query, key, and value vectors respectively. These attended image features $\phi_i^{'1} \in \mathbb{R}^{w_{I}h_{I} \times d}$ are first transposed and further processed as follows:

\begin{equation}
\label{eq:merge}
    \phi_i^1 = \mathbf{W}_2\left(ReLU\left(\mathbf{W}_1\phi_i^{'1}\right)\right) + \phi_i^1.
\end{equation}
The image features $\phi_i^{1}$ are reshaped to the original dimension before feeding them through the next block and the whole process is repeated for remaining blocks in the image encoder. The similarity scores between the target image and the sketch features are first calculated in equation~\ref{eq:soft} and these scores are then used to fuse relevant sketch features into the target image features as shown in equation~\ref{eq:merge} leading to better alignment between the two. Image features from each block of the sketch-guided vision transformer encoder are extracted and concatenated before passing them to the decoder as represented by $\phi_i^{[1,\dots,N]}$ in Figure~\ref{fig:archi}.

\subsubsection{(ii) Object and Query representation Refinement:}
\label{sec:prop_ref}
The decoder takes the image features at different scales to update the representation for the $[\mathtt{DET}]$ tokens. During training, the $[\mathtt{DET}]$ tokens are transformed into the representation of various objects at different locations in the image. Since it takes sketch-conditioned image representation at the input, the object features thus learned are better aligned with the query sketch. However, the query fusion that happened at the sketch-guided vision transformer encoder is at the coarse-grained image level. A more fine-grained fusion at the object level can lead to further improvement in performance.

Therefore, at the output of the decoder, the learned object features represented by the $[\mathtt{DET}]$ tokens and the sketch features are refined further to bring the representation of relevant objects closer to the query sketch for better scoring. Given the flattened sketch representation $\phi_s \in \mathbb{R}^{ wh \times d}$, the representations of $[\mathtt{DET}]$ tokens $\phi_{[\mathtt{DET}]} \in \mathbb{R}^{100 \times d}$ are refined using multi-headed cross-attention as follows:

\begin{equation}
\label{eq:sum}
    \phi_{[\mathtt{DET}]} = \phi_{[\mathtt{DET}]} + \mathbf{W}_3\left(ReLU\left(\mathbf{W}_4\phi_{[\mathtt{DET}]}^{'T}\right)\right), 
\end{equation}

\begin{equation}
\label{eq:weighing}
    \phi_{[\mathtt{DET}]}^{'} = softmax\left(\frac{(\phi_{[\mathtt{DET}]}\mathbf{W}_{Q_2})\left(\phi_s\mathbf{W}_{K_2}\right)^T}{\sqrt{d^{'}}}\right)\phi_s\mathbf{W}_{V_2},
\end{equation}
where $\mathbf{W}_{Q_2}, \mathbf{W}_{K_2}\in \mathbb{R}^{d\times d'}$ are the projection matrices for the query $\phi_{\mathtt{DET}}$ and the key $\phi_s$ respectively and $\mathbf{W}_{V_2} \in \mathbb{R}^{d\times d}$ is the projection matrix for the value $\phi_s$ in the attention calculation. 2D sinusoidal position encoding are added to the sketch representation before calculating the attention scores in equation~\ref{eq:weighing}. For the sake of brevity, attention computation is shown for a single head. Likewise, given the representation of $[\mathtt{DET}]$ tokens the representation of the query sketch are refined by using a separate cross-attention module similar to equation~\ref{eq:sum} and~\ref{eq:weighing}. Once the representation of the objects and the sketches are refined, they are then scored to localize all the instance of the corresponding object.

% \begin{equation}
%     \phi_s = \phi_s +  \mathbf{W}_5\left(ReLU\left(\mathbf{W}_6\phi_s^{'}\right)\right),
% \end{equation}

% \begin{equation}
%     \phi_s^{'} = softmax\left(\frac{(\phi_s\mathbf{W}_{Q_3})\left(\phi_{[\mathtt{DET}]}\mathbf{W}_{K_3}\right)}{\sqrt{d^{'}}}\right)\phi_{[\mathtt{DET}]}\mathbf{W}_{V_3},
% \end{equation}
% where $\mathbf{W}_{Q_3}, \mathbf{W}_{K_3}\in \mathbb{R}^{d\times d'}$ are the projection matrices for the query $\phi_s$ and the key $\phi_{\mathtt{DET}}$ respectively and $\mathbf{W}_{V_3} \in \mathbb{R}^{d\times d}$ is the projection matrix for the value $\phi_{\mathtt{DET}}$ in our attention framework.

\subsubsection{(iii) Scoring:}
\label{sec:prop_scoring}
At the output of the decoder, once we get the representation of the objects in the target image given by $[\mathtt{DET}]$ tokens, a scoring function $\Theta$ is learned to score each object with that of the query sketch. In order to do that, each of the $[\mathtt{DET}]$ tokens is assigned to each box in the ground truth using the Hungarian matching algorithm~\cite{DBLP:books/daglib/p/Kuhn10} as described in ~\cite{Carion2020EndtoEndOD}. Each token ($[\mathtt{DET}]_k$) is assigned a label $y_k$ (1 or 0) based on whether it is assigned to a bounding box containing a foreground object, i.e., the object that corresponds to the sketch query. A global representation for the sketch is then obtained by taking the max-pool of the sketch feature maps:
\begin{equation}
    \phi_S = \Psi(\phi_s),
\end{equation}
where $\phi_S \in \mathbb{R}^d$ and $\Psi: \mathbb{R}^{d\times w \times h} \rightarrow \mathbb{R}^d$.

Each of the $[\mathtt{DET}]$ token representations is concatenated with the global sketch representation before passing it through a neural network to generate the score for that token.

\begin{equation}
    score\left([\mathtt{DET}]_k, s\right) = \sigma\left( \Theta\left([\phi_{[\mathtt{DET}]_k}; \phi_S]\right)\right),
\end{equation}
where $\Theta$ is a neural network, $\sigma$ is sigmoid function and $score:\mathbb{R}^{2d}\rightarrow [0,1]$. The model is then trained to give high scores to the tokens which correspond to the objects in the query sketch by minimizing the following loss function:

% \begin{equation}
% \label{eq:loss}
%     L([\mathtt{DET}], s) = \sum_{k} \left\{y_k score([\mathtt{DET_k}], s) + (1-y_k)(1-score([\mathtt{DET_k}],s))\right\}
% \end{equation}

\begin{multline}
    \label{eq:loss}
    L([\mathtt{DET}], s) = \sum_{k} \Bigl\{-y_k \ln\left(score([\mathtt{DET}]_k, s)\right) - \\
    (1-y_k)(1-\ln\left(score([\mathtt{DET}]_k,s))\right)\Bigr\}.
\end{multline}
Along with the classification loss defined in equation~\ref{eq:loss}, regression loss and Generalized IoU~\cite{DBLP:conf/cvpr/RezatofighiTGS019} loss is also defined on the predicted bounding boxes with respect to the ground truth bounding box.

During inference, all the $[\mathtt{DET}]$ tokens are scored with the query sketch and the bounding boxes corresponding to high scoring tokens are selected as the localized objects. 
\begin{table*}[h]
\centering
\begin{tabular}{l|rr|rr}
\toprule
\multirow{2}{*}{Model} & \multicolumn{2}{|c}{Query: Sketchy}                                          & \multicolumn{2}{|c}{Query: QuickDraw}                                        \\
                       & mAP (\%)  & AP@50 (\%)  & mAP (\%)  & AP@50 (\%) \\
\midrule
Detection-based*        &      &              &                       &                       \\
~~~~~~~~FasterRCNN~\cite{Ren2015FasterRT}             & 40.2 & 64.4                       & 35.5 & 58.1  \\
~~~~~~~~Retinanet~\cite{DBLP:conf/iccv/LinGGHD17}              & 42.2 & 66.1                       & 37.9 & 60.1  \\
~~~~~~~~DETR~\cite{Carion2020EndtoEndOD} &47.0& 68.7& 41.1&62.7\\
Localization-based        &      &              &                       &    \\ 
~~~~~~~~Modified FasterRCNN~\cite{Ren2015FasterRT}    & -    & -                          & 18.2 & 31.5 \\
~~~~~~~~CoAT~\cite{Hsieh2019OneShotOD}                   & -    & -                          & 27.9 & 48.6  \\
~~~~~~~~CMA~\cite{Tripathi2020SketchGuidedOL}  & -    & -                         & 30.0 & 50.0  \\
~~~~~~~~Sketch-DETR~\cite{Riba2021LocalizingIF}            & 42.0 & 63.6                     & 41.4 & 62.1  \\
~~~~~~~~\textbf{Ours}                  & \textbf{50.0} (8.0 $\uparrow$) & \textbf{73.9} (10.3 $\uparrow$)  & \textbf{48.0} (6.5 $\uparrow$) & \textbf{71.7} (9.6 $\uparrow$)\\
\bottomrule
\end{tabular}
\caption{\label{tab:common_one} Results in \textbf{close-world, one-shot} setting. During inference, single sketch from Sketchy and QuickDraw! respectively has been used as a query to localize 'seen' object categories on target images from \textit{MS-COCO val2017} dataset, and mean average precision and AP@50 computed over all sketch queries have been reported. The numbers inside parenthesis show gain with respect to the most competitive localization-based baseline. *: Detection-based baselines assume availability of set of object categories.}
\end{table*}

\subsection{Multi-query localization}\label{sec:mutli_query} The sketches in the dataset are generally abstract and contain minimal information about the objects' shapes and attributes. However, as noted in~\cite{Tripathi2020SketchGuidedOL} using multiple sketch queries for localization may provide complementary information that can help to improve the localization of the corresponding objects in the natural image. Although our method is effective even with a single sketch query (one-shot), in order to utilize the information from the multiple sketches and perform multi-query localization in our localization framework, we first modify Equation~\ref{eq:soft} and Equation~\ref{eq:merge} as follows:
\begin{equation}
    \phi_i^{'1}[l] = softmax\left(\frac{(\phi_i^1\mathbf{W}_{Q_1})\left(\phi_{s_l} \mathbf{W}_{K_{1}}\right)^T}{\sqrt{d^{'}}}\right)\phi_{s_l}\mathbf{W}_{V_1},
\end{equation}
where, $\phi_i^{'1}[l]$ is the attention aggregated features for the $l^{th}$ sketch query. These features are aggregated for each query sketch, transposed, and then added to the image representations as follows:
\begin{equation}
    \phi_i^1 = \mathbf{W}_2\left(ReLU\left(\frac{1}{L}\sum_{L=1}^L \mathbf{W}_1\phi_i^{'1}[l]\right)\right) + \phi_i^1,
\end{equation}
where $L$ is the total number of query sketches. 

Further at the decoder,  we propose an attention-based query fusion strategy to construct a unified sketch query representation from multiple sketches. Given $n$ sketches and their feature map representations, we first take the average across the sketches to get the averaged feature map representation denoted as $\phi_{s_{\mu}} \in \mathbb{R}^{d\times w\times h}$. Each of the query sketch representations is flattened and concatenated and is represented as $\phi_{s_{\{1,L\}}}$. We then utilize attention to incorporate complementary information present among the diverse sketches into the average sketch representation. The attention-based query fusion is defined as follows:

\begin{equation}
    \phi_{s_{\mu}} = \phi_{s_{\mu}} + \mathbf{W}_5\left(ReLU\left(\mathbf{W}_6\phi_{s_{\mu}}^{'T}\right)\right),
\end{equation}

\begin{equation}
    \phi_{s_{\mu}}^{'} = softmax\left(\frac{\left(\phi_{s_{\mu}}\mathbf{W}_{Q_3}\right)\left(\phi_{s_{\{1,L\}}} \mathbf{W}_{K_{3}}\right)^T}{\sqrt{d^{'}}}\right)\phi_{s_{\{1,L\}}}\mathbf{W}_{V_{3}},
\end{equation}
where $\mathbf{W}_{Q_3}, \mathbf{W}_{K_3}\in \mathbb{R}^{d\times d'}$ are the projection matrices for the query $\phi_{s_{\mu}}$ and the key $\phi_{s_{\{1,L\}}}$ respectively and $\mathbf{W}_{V_3} \in \mathbb{R}^{d\times d}$ is the projection matrix for the value $\phi_{s_{\{1,L\}}}$.
% Moreover, $\mathbf{W}_1, \mathbf{W}_2, \mathbf{W}_3, \mathbf{W}_4, \mathbf{W}_5, \mathbf{W}_6, \mathbf{W}_7, \mathbf{W}_8 \in \mathbb{R}^{d\times d}$ are the learned projection matrices.

The proposed method first learns the correspondences between the average query feature and all the query features and then use these correspondences to fuse complementary information present in diverse sketches into the average sketch representation. This fused sketch representation is then used as the query in the refinement and the scoring stage. Moreover, $\mathbf{W}_i$ where $i=\{1,\cdots, 6\}$ are the learned projection matrices.

\section{Experiments and Results}
\subsection{Datasets and Evaluation Setup:} In this work, we use images from the MS-COCO dataset as target scenes and sketches from the QuickDraw! and Sketchy datasets as queries to evaluate the performance of our model. Sketchy~\cite{sangkloy2016sketchy} contains 75,471 samples across 125 object categories. For each image, a sketch is drawn by the crowd worker. Therefore, there is a fine-grained association between the images and sketches in this dataset. QuickDraw!~\cite{Ha2018ANR} has 50M drawings for 345 object categories. The sketches in this dataset are available as vector drawings, and we rasterize them before feeding them into the sketch encoders.
For target scenes, we use
MS-COCO~\cite{Lin2014MicrosoftCC}. There are 56 and 27 object categories common between MS-COCO and QuickDraw! and MS-COCO and Sketchy, respectively. We used the images with the common objects from the COCO train2017 dataset and evaluated them on COCO val2017 dataset.

We evaluate the performance of the proposed model using the following two setups: (i) open-world one-shot and (ii) close-world one-shot. A single sketch is used to query the target image in the one-shot setup. In the open-world setting, 14 categories out of 56 common categories between QuickDraw! and MS-COCO are removed from the training data and are called `unseen' categories. The data corresponding to the `unseen' categories is also removed from the Imagenet during pretraining to ensure the correct open-world setting. Similarly, the sketch encoder is pretrained on QuickDraw! dataset after removing all the aforementioned categories corresponding to the `unseen' classes. Additionally, we also perform experiments in the multi-query setup where a set of five sketch queries are used.
\begin{table}[!t]
% \small
\centering
\begin{tabular}{lrrr}
\toprule
Models                & mAP & AP@50 & AP$^L$ \\
\midrule
Modified FasterRCNN & 3.3 & 7.4& 6.2   \\
CoAT~\cite{Hsieh2019OneShotOD}                  & 5.9 & 12.4  & 10.6  \\
CMA~\cite{Tripathi2020SketchGuidedOL} & 7.5 & 15.0  &  12.4  \\
\textbf{Ours}                  & \textbf{12.2} & \textbf{18.3}  & \textbf{24.6} \\
\bottomrule
\end{tabular}
\caption{\label{tab:disjoint_one} Results in \textbf{open-world, one-shot} setting. Here, single sketch query from QuickDraw! has been used to perform localization on `unseen' categories of \textit{COCO val2017} dataset and mean performance over all sketch queries has been reported. We observe that for large-sized object categories for example, \emph{elephant}, \emph{bear}, \emph{bus}, etc.  our approach outperforms state-of-the-art published result by 12.2\% as measured by AP$^L$.}
\end{table}
\begin{table}[!t]
\small
\centering
\begin{tabular}{l|rr|rr}
\toprule
\multirow{2}{*}{Model} & \multicolumn{2}{|c}{Sketchy} & \multicolumn{2}{|c}{QuickDraw} \\
                       & mAP          & AP@50        & mAP           & AP@50         \\
\midrule
CMA  & -            & -            & 30.0          & 50.0          \\
~~+ Query Fusion (5Q)     & -            & -            & 32.0          & 52.6          \\
~~+ Feature Fusion (5Q)   & -            & -            & 32.0          & 53.1          \\
Ours                  & 50.0         & 73.9         & 48.0          & 71.7          \\
~~+ Attention Fusion (5Q)     & \textbf{50.7}         & \textbf{74.7}         &  \textbf{49.2}        &     \textbf{72.6} \\
\bottomrule
\end{tabular}
\caption{\label{tab:multi_common}Results in \textbf{multi-query} setting. Here five sketch queries (5Q) from the QuickDraw! are used to query images from \textit{COCO val2017} dataset.}
\end{table}

\subsection{Baselines} We used the following baselines in our experiments:
\subsubsection{Detection-based baselines:} In these baselines, the target image is passed through the object detectors to predict a set of bounding boxes along with the corresponding classes, and the query sketch is passed through a sketch classifier to predict the class for the sketch. Then the predicted category of the sketch is used to get the corresponding localizations from the predictions of the object detectors. We use FasterRCNN~\cite{Ren2015FasterRT}, Retinanet~\cite{Lin2020FocalLF}, and DETR~\cite{Carion2020EndtoEndOD} for comparison. It should be noted here that these baselines assume the prior availability of the set of object categories and, therefore, can only be evaluated in close-world setting. 
\subsubsection{Localization-based baselines:} In these baselines, the sketch queries are directly compared with the representation of objects in the image to generate the localizations. We used the following baselines in our experiments: (i) \textbf{Modified FasterRCNN}: The region proposals are first generated using FasterRCNN. Then the representation of region proposals is scored with the query sketch representation to obtain the localizations. \textbf{(ii) Query-guided RPN}: In these baselines, the query information is incorporated in a region proposal network (RPN) to generate the region proposals relevant to the sketch query. To this end, we compared against two recent techniques namely, CoATex~\cite{Hsieh2019OneShotOD} and cross-modal attention (CMA)~\cite{Tripathi2020SketchGuidedOL}. \textbf{(iii) Sketch-DETR~\cite{Riba2021LocalizingIF}:} uses a transformer-based object detector and concatenates the sketch query tokens with the image tokens at the input of the DETR encoder to incorporate the query information. This method has shown state-of-the-art results in sketch-based object localization. 

%\subsubsection{Implementation Details}
%The proposed model was implemented using the Pytorch v1.9.1~\cite{Paszke2019PyTorchAI} with CUDA 11.1. The model is trained on a batch size of 7 on a single Nvidia-Quadro 8000 GPU with  stochastic gradient descent (SGD) with the momentum of $0.9$. The model is end-to-end trained for 14 epochs with starting learning rate of $1e-5$ and a factor of $0.1$ decays the learning rate after $8$ epochs. The implementation of the model is provided in the supplementary materials.

\begin{table}[!t]
\centering
\begin{tabular}{lrr}
\toprule
Model                 & mAP  & AP@50 \\
\midrule
Vanilla ViDT                  & 39.4 & 56.6  \\
~~~~ + Sketch-guided Vis. Trans. & 46.9 & 68.7  \\
~~~~ + Obj. and Query Refinement       & \textbf{48.0} & \textbf{71.7} \\
\bottomrule
\end{tabular}
\caption{\label{tab:ablation}Effect of various components on the performance of the proposed model. The results are reported for images from COCO \textit{val2017} and queries from QuickDraw! dataset.}
\end{table}

\begin{figure*}[!t]
    \centering
    \includegraphics[width=0.94\textwidth]{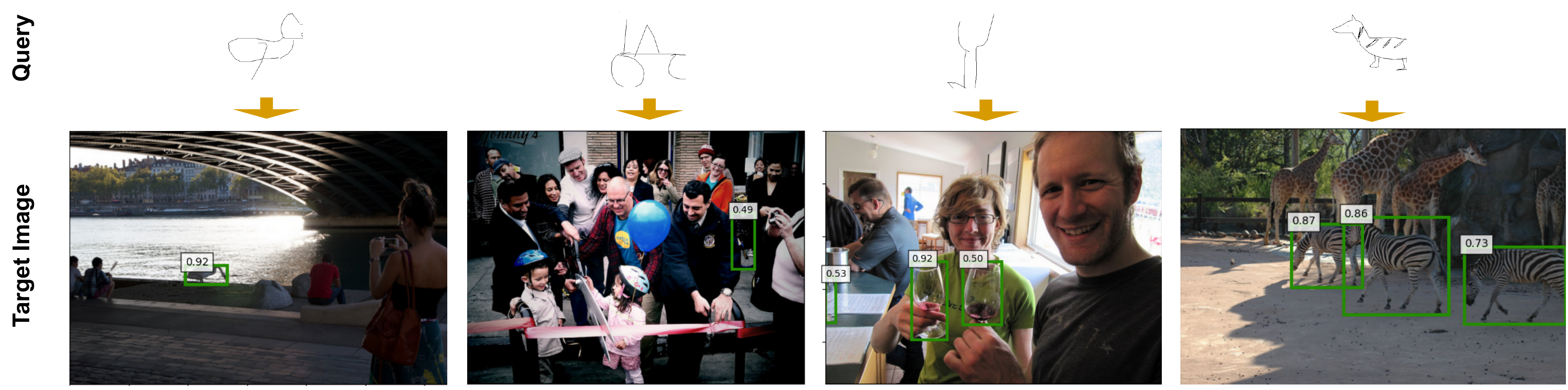}
    \caption{\label{fig:visRes}\textbf{Qualitative Results:} A selection of sketch queries and target images with queried object localized using our proposed model are shown in first and second row respectively. Our proposed model is able to localize occluded object (such as \emph{bicycle} in third column) as well as multiple object instance (such as \emph{glass} and \emph{zebra} in third and fourth column) successfully. \textbf{[Best viewed in color]}}
\end{figure*}

\subsection{Results and Discussion}
\subsubsection{Close-world one-shot localization:}
We first show the results of sketch-based object localization for close-world, one-shot setting in Table~\ref{tab:common_one}. The proposed model outperforms the state-of-the-art by a large margin. The Modified Faster RCNN performs poorly because of the query-independent region proposal network. Although the query-dependent RPNs used in CoAT and Cross-Modal Attention lead to significant improvement in performance, they fall short compared to the detection-based methods and Sketch-DETR. The detection-based methods try to cover the domain gap present in images and sketches by doing object detection and sketch recognition separately and then later mapping the predicted category of the sketch to the detected objects on the image. However, these methods are limited by the performance of the object detectors and sketch classifiers and require a set of object categories to be known a priori. Sketch-DETR, on the other hand, uses a DETR-based localization framework; however, the alignment between the sketch and image features is limited using this method, leading to weak performance. In contrast, the proposed sketch-guided vision transformer encoder leads to query-aligned image features. Further, with object-level refinement at the output of the decoder, the proposed model leads to state-of-the-art localization performance.

\subsubsection{Open-world one-shot localization:}
The localization results in this challenging setting are provided for the MS-COCO val2017 dataset in Table~\ref{tab:disjoint_one}. In this setting, the proposed models outperform the current state-of-the-art method by $1.6\%$ mAP and $7.2\%$ improvement in AP@50 for large objects. Modified FasterRCNN utilizes a vanilla RPN to generate region proposals for `unseen' objects, which perform poorly. On the other hand, cross-modal attention gives performance improvement by utilizing a query-dependent region proposal network. The proposed model utilizes a better image and sketches alignment strategy and learns a generalized query-conditioned feature. This enables improved performance even for unseen objects. Especially in the case of large objects, the performance gain as compared to best-reported results is significant ($\approx  7\%$). 

\subsubsection{Multi-query localization:}
We now present the effect of the proposed attention-based query fusion strategy, described in section~\ref{sec:mutli_query} for multi-query object localization. The results are shown in Table~\ref{tab:multi_common} for target images from the MS-COCO dataset and queries from the QuickDraw! and the Sketchy datasets. In this experiment, we used five sketches to query the target image. The proposed fusion strategy can fuse complementary information among different query sketches and improve localization performance. 

%Moreover, to evaluate the robustness of the proposed query fusion strategy to the number of query sketches, we used a localization model trained on five sketch queries and evaluated the model on two to eight sketch queries. The localization performance of the model measure in mAP is given by 50.5\pm0.11 for the Sketchy dataset. The model's performance does not vary significantly with the number of query sketches, establishing the robustness of the proposed fusion strategy.

% The results are shown in Table~\ref{tab:num_sk} for the sketch queries from the Sketchy and QuickDraw! datasets. The model's performance does not vary significantly with the number of query sketches, establishing the robustness of the proposed fusion strategy.
\begin{table}[t]
\centering
\begin{tabular}{lcc}
\toprule
Model         & mAP  & AP@50 \\
\midrule
Modified-ViDT & 39.4 & 56.6  \\
CMA-ViDT      & 42.3 & 63.5  \\
Sketch-ViDT   & 43.0 & 66.8  \\
\midrule
Ours (w/o OQR)       & \textbf{46.9} & \textbf{68.7}  \\
\bottomrule
\end{tabular}
\caption{\label{tab:atten_comparison} Comparison of the proposed sketch-guided vision transformer encoder with the late fusion strategies proposed in CMA and Sketch-DETR. The results are reported for queries from QuickDraw! dataset. Here, OQR refers to proposed object and query refinement.}
\end{table}
\subsubsection{Ablation:}
We now describe the effect of each component on the performance of the proposed model. The results are shown in Table~\ref{tab:ablation}. The query-aligned image features obtained at the output of the sketch-guided vision transformer encoder gives the most performance improvement. Moreover, the object-level feature refinement at the output of the decoder further improves the localization performance.

\subsubsection{Comparison of feature fusion methods:}
The cross-modal attention proposed in~\cite{Tripathi2020SketchGuidedOL} and multi-headed attention used in~\cite{Riba2021LocalizingIF} can be thought of as late-fusion of the query sketch information in the target image features. We incorporated the cross-modal and multi-headed attention into the ViDT architecture, compared it with the proposed early fusion method in the sketch-guided vision transformer encoder, and reported the results in Table~\ref{tab:atten_comparison}. The superior performance of the proposed early fusion suggests that early fusion leads to better alignment between the target image and the query sketch features leading to better localization. 

\subsubsection{Qualitative Results:} We perform a detailed qualitative analysis of our approach. A selection of results are shown in Figure~\ref{fig:visRes}. We observe that our method is successful in localizing occluded as well as multiple instance of objects. A detailed analysis is shown in the supplementary material.

\section{Conclusion}
In this work, we extensively studied the problem of sketch-guided object localization in natural images and proposed a novel transformer-based end-to-end trainable model. The proposed model uses the novel sketch-guided vision transformer encoder to learn sketch-conditioned image features. Further, object-level feature refinement at the output of the decoder is performed to align the natural image and the query sketch features effectively. The effectiveness of the proposed model has been established by the state-of-the-art performance on publicly available benchmarks. Nevertheless, despite significant improvement in localization performance, sketch-based object localization is yet to be ready for deployment, inviting more research in this direction. We firmly believe that this work shall trigger more research efforts toward solving this important and challenging problem.   

%The proposed model can naturally localize the `unseen' objects because of the query-aligned image features that are learned in our model. Moreover, we extended our framework to enable querying using multiple sketches and proposed an attention-based query fusion strategy that combines the complementary information among many queries to create a complete representation of the object leading to better localization performance.  
% Use \bibliography{yourbibfile} instead or the References section will not appear in your paper
\bibliography{aaai23}

\end{document}